\documentclass[sigconf]{acmart}

\usepackage{multirow}
\usepackage{multicol}
\usepackage{graphicx}
\usepackage{adjustbox}
\usepackage{array}
\usepackage{bbm}
\usepackage{amsfonts, amsmath}
\usepackage{xcolor,colortbl}
\usepackage[ruled,vlined,linesnumbered,boxed]{algorithm2e}
\usepackage{algpseudocode}
\usepackage{booktabs}

\usepackage{xspace}

\usepackage{subfigure}

\usepackage{tcolorbox}
\usepackage{colortbl}

\usepackage{enumitem}

\definecolor{LightCyan}{rgb}{0.88,1,0.88}
\definecolor{LightRed}{rgb}{1,0.88,0.88}
\definecolor{beaublue}{rgb}{0.9, 0.95, 0.9}
\definecolor{blackish}{rgb}{0.2, 0.2, 0.2}

\usepackage{soul}

\definecolor{LightCyan}{rgb}{0.88,1,1}
\sethlcolor{LightCyan}

\makeatletter
\def\SOUL@hlpreamble{%
    \setul{}{2.5ex}
    \let\SOUL@stcolor\SOUL@hlcolor
    \dimen@\SOUL@ulthickness
    \dimen@i=-.75ex 
    \advance\dimen@i-.5\dimen@
    \edef\SOUL@uldepth{\the\dimen@i}%
    \let\SOUL@ulcolor\SOUL@stcolor
    \SOUL@ulpreamble
}
\makeatother

\newcommand*{\codebox}[1]{{\hl{#1}}}

\makeatletter
\DeclareRobustCommand\onedot{\futurelet\@let@token\bmv@onedotaux}
\def\bmv@onedotaux{\ifx\@let@token.\else.\null\fi\xspace}
\def\eg{\emph{e.g}\onedot} 
\def\ie{\emph{i.e}\onedot} 
\def\cf{\emph{c.f}\onedot} 
 \def\vs{\emph{vs}\onedot}
\def\wrt{w.r.t\onedot} 
\def\etal{\emph{et al}\onedot}

\makeatother

\makeatletter
\newcommand\footnoteref[1]{\protected@xdef\@thefnmark{\ref{#1}}\@footnotemark}
\makeatother

\AtBeginDocument{%
  \providecommand\BibTeX{{%
    \normalfont B\kern-0.5em{\scshape i\kern-0.25em b}\kern-0.8em\TeX}}}

\AtBeginDocument{%
  \providecommand\BibTeX{{%
    Bib\TeX}}}

\setcopyright{acmlicensed}
\copyrightyear{2018}
\acmYear{2018}
\acmDOI{XXXXXXX.XXXXXXX}

\acmConference[WSDM '25]{The 18th ACM International Conference on Web Search and Data Mining}{March 10th-14th,
  2025}{Hannover, Germany}
\acmISBN{978-1-4503-XXXX-X/18/06}

\begin{document}

\title{Inductive Graph Few-shot Class Incremental Learning}



\author{Yayong Li}
\authornote{Most work done at CSIRO.}
\email{yayong.li@uq.edu.au}
\affiliation{
  \institution{The University of Queensland}
  \city{Brisbane}
  \country{Australia}
}

\author{Peyman Moghadam}
\email{peyman.moghadam@csiro.au}
\affiliation{
  \institution{CSIRO Robotics, CSIRO}
  \city{Brisbane}
  \country{Australia}
}

\author{Can Peng}
\authornotemark[1]
\email{can.peng@eng.ox.ac.uk}
\affiliation{
  \institution{University of Oxford}
  \city{Oxford}
  \country{United Kingdom}
}

\author{Nan Ye}
\email{nan.ye@uq.edu.au}
\affiliation{
  \institution{The University of Queensland}
  \city{Brisbane}
  \country{Australia}
}

\author{Piotr Koniusz}
\authornote{Corresponding author. $\quad$ This work is accepted by the 18th ACM International Conference on Web Search and Data Mining (WSDM), 2025.}
\email{piotr.koniusz@csiro.au}
\affiliation{
  \institution{Data61, CSIRO}
  \city{Canberra}
  \country{Australia}
}


\renewcommand*{\thefootnote}{\arabic{footnote}}
\begin{abstract}
Node classification with Graph Neural Networks (GNN) under a fixed set of labels is well known in contrast to Graph Few-Shot Class Incremental Learning (GFSCIL), which involves learning a GNN classifier as graph nodes and classes growing over time sporadically. We introduce \textit{inductive GFSCIL} that continually learns novel classes with newly emerging nodes while maintaining performance on old classes without accessing previous data. This addresses the practical concern of transductive GFSCIL, which requires storing the entire graph with historical data\footnote{\label{note1}A well-established research on incremental learning in computer vision \cite{tao2020few, zhang2023few} does not permit access to the whole of the past training data due to privacy preservation and memory limitation considerations. By analogy, graph incremental learning models should not store the subgraphs corresponding to the past sessions, making such a setting an inductive GFSCIL setting.}. Compared to the transductive GFSCIL, the inductive setting exacerbates catastrophic forgetting due to inaccessible previous data during incremental training, in addition to overfitting issue caused by label sparsity. Thus, we propose a novel method, called \textbf{T}opology-based class \textbf{A}ugmentation and \textbf{P}rototype calibration (\textbf{TAP}). To be specific, it first creates a triple-branch multi-topology class augmentation method to enhance model generalization ability. As each incremental session receives a disjoint subgraph with nodes of novel classes, the multi-topology class augmentation method helps replicate such a setting in the base session to boost backbone versatility. In incremental learning, given the limited number of novel class samples, we propose an iterative prototype calibration to improve the separation of class prototypes. Furthermore, as backbone fine-tuning poses the feature distribution drift, prototypes of old classes start failing over time, we propose the prototype shift method for old classes to compensate for the drift. We showcase the proposed method on four datasets. 

\end{abstract}





\maketitle

\section{Introduction}
\begin{figure}[t]
\vspace{-0.3cm}
    \centering
    \subfigure[Transductive GFSCIL]{
    \includegraphics[trim={0 0 0 19},clip,width=0.4\textwidth]{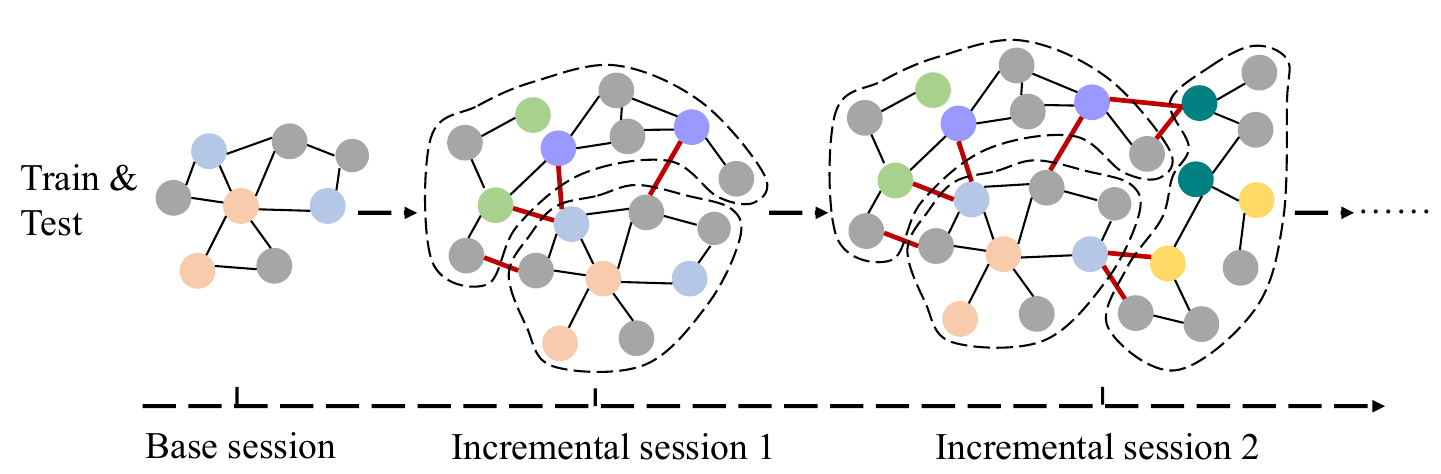}
    \label{fig:transductive}
    }
    \subfigure[Inductive GFSCIL]{
    \includegraphics[trim={0 0 0 0},clip,width=0.4\textwidth]{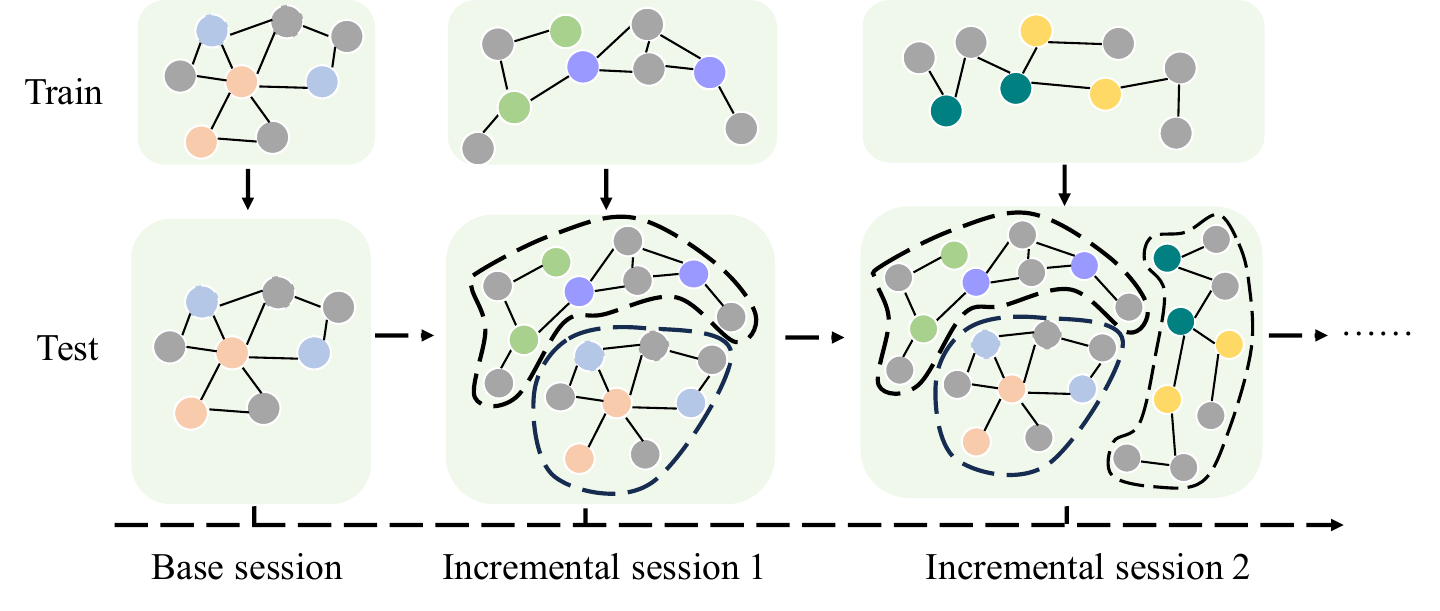}
    \label{fig:inductive}
    }
    \vspace{-0.1cm}
    \caption{Transductive GFSCIL in Fig. \ref{fig:transductive} \vs. our inductive GFSCIL in Fig. \ref{fig:inductive}. The classes are distinguished by distinct colors, with grey indicating unlabeled nodes for testing. Graphs that emerged at distinct sessions are enclosed by dashed lines. Notice that in transductive GFSCIL, the graph gets larger with each session, and links among session subgraphs are required (red color). In contrast, inductive GFSCIL does not store past subgraphs.}
    \vspace{-0.1cm}
\end{figure}

Node classification via Graph Neural Network (GNN) \cite{kipf2016semi, hamilton2017inductive, velivckovic2017graph, hu2020open} is a well-studied topic. However, most existing GNNs operate within a predefined class space in contrast to recent trends in lifelong learning that permit open-ended label space \cite{ROY202365}. As time progresses, the graph evolves and novel categories emerge with small quantities of additional nodes and links that appear over time. Thus, GNNs should 
have the ability to be incrementally updated to capture novel classes without the performance loss on old classes. 
Recently proposed Graph Few-Shot Class Incremental Learning (GFSCIL) \cite{tan2022graph, lu2022geometer} primarily concentrated on the transductive setting. As depicted in Figure~\ref{fig:transductive}, new nodes and links that emerge with time are appended to the graph of the previous session. Hence, the model has access to the entire graph. 

However, this assumption does not always align with real-world scenarios. In practice, it is common for newly formed (sub)graphs to arise and exist autonomously, with nodes of novel classes devoid of any connections with antecedent graphs.  
This phenomenon is particularly conspicuous in social media networks, such as Reddit or Twitter(X), where rapidly changing events frequently foster the formation of closely-knit communities and groups centered around various trending topics over time. Due to the prohibitive memory and computational demands, individuals often process subgraph snapshots sequentially within specific timeframes, instead of using the whole giant graph. These snapshots naturally exhibit high independence, each associated with distinct trending topics, without establishing connections to pre-existing subgraph snapshots. Consequently, the resulting subgraphs are self-contained and disjoint from previous graphs. Moreover, due to considerations such as privacy preservation, memory limitations, or other practical constraints, (\eg archived posts) ~\cite{tao2020few, zhang2023few}, historical data samples are typically unavailable during new learning sessions, highlighting the need to explore inductive GFSCIL. Furthermore, the transductive GFSCIL also stands in contrast to the well-established incremental learning scenarios in computer vision \cite{tao2020few, zhang2023few} where one is prohibited from making use of the past training data.

Thus, in this paper, we propose and investigate \textit{\textbf{inductive}} GFSCIL depicted in Figure \ref{fig:inductive}. 
Generally, the model is initially trained with one base session containing a large portion of data in order to pre-train well the backbone. Subsequently, multiple small-scale incremental sessions are used for fine-tuning. 
In each incremental session, $N$ previously unseen classes with $K$ labeled nodes (samples) per class are given, and $Q$ unlabeled nodes\footnote{The value of $Q$ may vary from session to session.}. The task is to update the model by $NK$ labeled and $Q$ unlabeled nodes from the unseen subgraph and then perform classification on both old and novel classes.
Intuitively, the inductive setting assumes that the newly emergent subgraph is semantically homogeneous and topologically independent of subgraphs of the past sessions. Its node set, link set, and label space are disjoint with antecedent graphs. However, the node attribute space is shared and one hopes that topological patterns among subgraphs share some similarities. Moreover, the antecedent graphs are entirely unavailable for training.
Unlike transductive GFSCIL, which primarily tackles class imbalance, inductive GFSCIL tackles the stability-plasticity dilemma, akin to typical FSCIL in computer vision \cite{zhu2021self, cheraghian2021semantic, chi2022metafscil}. 

The stability-plasticity dilemma describes the challenge of gaining the ability to classify novel classes and preserving the ability to recognize old classes. Due to the inaccessibility of preceding subgraphs, fine-tuning the model on the novel subgraph might erase the model's memory (the so-called catastrophic forgetting), compromising its recognition ability for old classes. Moreover, preserving the recognition ability for old classes by limiting the model adaption sacrifices its plasticity--the model will fail to learn/recognize novel classes.  
Finally, the lack of a large amount of labeled nodes of novel classes may lead to overfitting in the incremental learning steps. 


To cope with the aforementioned issues, we propose the \textbf{Topology-based class Augmentation and Prototype calibration (TAP)} method for the inductive GFSCIL problem.  
For the base session, we propose a triple-branch multi-topology class augmentation designed to preserve space for novel classes, thereby enhancing the model’s generalization capacity. This approach allows the model to simulate possible novel semantic and topological patterns, \eg, by dividing label space into subsets and the base graph into corresponding disjoint subgraphs, emulating the disjoint nature of incremental sessions. In the incremental session, the model undergoes fine-tuning for a few epochs to integrate the knowledge of novel classes, with an exponential moving average applied then to the past and current session models to mitigate catastrophic forgetting.
Concerning the potential bias arising from limited novel samples, an iterative prototype calibration method has been proposed to improve the representativeness of novel prototypes. 
Furthermore, since fine-tuning the model results in feature distribution drift—leading to deviation with older class prototypes—we introduce a prototype shift method to restore the representativeness of these older prototypes. Our contributions are summarized below:
%

\vspace{-0.1cm}
\renewcommand{\labelenumi}{\roman{enumi}.}
\begin{enumerate}[leftmargin=0.6cm]
    \item We formulate the inductive GFSCIL setting for the node classification. By contrast to transductive GFSCIL, the inductive setting deals with a more challenging practical scenario of incremental learning without accessing the past graph data. 
    \item To alleviate the stability-plasticity dilemma, we propose a triple-branch multi-topology class augmentation, an iterative prototype calibration, and a prototype shift update to enable incremental learning of novel classes without compromising the performance on old classes.  
    \item We demonstrate the effectiveness of the proposed method on four datasets in comparison to state-of-the-art baselines.
\end{enumerate}





\section{Related Works}

\textbf{Few-shot Class Incremental Learning (FSCIL).} 
FSCIL, recently introduced in the context of image classification, learns new tasks with limited data while retaining high accuracy on old tasks without accessing the past data.
Tao \etal \cite{tao2020few} proposed a neural gas network that preserves the topology of the feature manifold for both old and new classes.
Subsequently, Zhao \etal \cite{zhao2021mgsvf} proposed a frequency-aware distillation 
while Zhang \etal \cite{zhang2021few} applied a graph attention network to adjust the final layer parameters to boost the accuracy.
Zhu \etal \cite{zhu2021self} developed a dynamic relation projection module to refine prototypes and improve class separation.

\vspace{0.1cm}
\noindent
\textbf{Graph Few-shot Class Incremental Learning (GFSCIL).} In 2022, FSCIL has been extended to 
node classification with a continually evolving graph. 
However, as new nodes emerge and expand the class space, they also affect the graph topology. Thus, simply applying existing computer vision FSCIL models to graphs fail \cite{zhu2021self, cheraghian2021semantic, chi2022metafscil, PENGtrips}. Thus, 
Tan \etal \cite{tan2022graph} proposed a hierarchical attention with pseudo-incremental training to estimate the importance of tasks. 
Lu \etal \cite{ lu2022geometer} enhanced the separation of classes by maintaining the geometric class relationships. 
However, both these works 
assume the transductive scenario to maintain access to nodes of past sessions. 
Such a practical limitation inspires us to develop inductive GFSCIL which does not require storing past nodes or continuously enlarging the graph.

\vspace{0.1cm}
\noindent
\textbf{Graph Class Incremental Learning (GCIL).} Transductive GFSCIL is similar to The GCIL, except that the latter models enjoy a sufficient number of labeled samples for incremental learning to avoid overfitting. 
To address catastrophic forgetting,  
Liu \etal \cite{liu2021overcoming} proposed a topology-aware weight-preserving method to slow down the update of pivotal parameters in incremental sessions. 
Zhang \etal \cite{zhang2022hierarchical} proposed a hierarchical prototypical network, which adaptively selects different feature extractors and prototypes for different tasks.
Wang \etal \cite{wang2022lifelong} developed a graph classification model in contrast to node classification. 
Zhou \etal \cite{zhou2021overcoming} and Liu \etal \cite{liu2023cat} proposed to store and replay important nodes to combat the forgetting problem.
However, the above methods are not applicable to inductive GFSCIL as they require a large number of samples in incremental sessions (\cf limited samples in GFSCIL) to avoid overfitting. GCIL is also predominantly transductive.


\vspace{0.1cm}
\noindent
\textbf{Graph Few-shot Learning (GFSL).} 
Similarly to GFSCIL, the GFSL approaches generalize to novel tasks with limited labeled samples. However, they do not handle the aspect of incrementally enlarging the label space. 
To facilitate adaptation to novel tasks, 
Wu \etal\cite{wu2022information} proposed a dual augmentation with self-training and noise injection.   
Liu \etal~\cite{liu2022few} proposed a meta-learning framework, using scaling and shifting transformations to improve the model transferability.
Wang \etal~\cite{wang2022task} proposed a triple-level adaptation module to alleviate the variance of different meta-tasks. 
Tan \etal~\cite{tan2022supervised} and Zhou \etal~\cite{zhou2022task} proposed contrastive learning for few-shot node classification.
Zhang \etal~\cite{zhang2022mul} exploited multi-level node relations to enable more transferrable node embeddings. Other contrastive problems include traffic \cite{scpt}, generalized Laplacian eigenmaps \cite{glen}, and preventing the dimensional collapse in the Euclidean \cite{decorell} and hyperbolic spaces \cite{outershellisotropy}. 


\section{Preliminaries}

\subsection{Problem statement}

Let  $\mathcal{D}=\{\mathcal{D}^0, \mathcal{D}^1, ..., \mathcal{D}^{m-1}\}$ be
 a sequence of $m$ homogeneous graph subsets of data, where $\mathcal{D}^i=\{\mathcal{G}^i, \mathcal{C}^i\}$. Let $\mathcal{G}^i=\{\mathcal{X}^i, \mathcal{A}^i\}$ be a graph with node features $\mathcal{X}^i$ and the adjacent matrix $\mathcal{A}^i$. Let $\mathcal{C}^i$ be the class space of $\mathcal{G}^i$.
Assume any two subsets of data, $\mathcal{D}^i, \mathcal{D}^j,i\!\neq\!j,$ have a disjoint label and graph space, \ie, $\mathcal{C}^i \cap \mathcal{C}^j\!=\!\varnothing $ and $\mathcal{G}^i \cap \mathcal{G}^j\!=\!\varnothing, \forall i\!\neq\!j$. 
Then, the model is trained on the sequential dataset $\mathcal{D}$ with $m$ separate learning sessions. We assume that $\mathcal{D}^0$ has a sufficient number of labeled nodes for training in the base session to obtain a well pre-trained backbone. Subsequently, we proceed with the incremental learning session $t$. Each  $\mathcal{D}^t,t>0,$ contains $N\!=\!|\mathcal{C}^t|$ novel classes\footnote{For brevity, assume incremental sessions have the same number of novel classes, \ie, $N\!=\!|\mathcal{C}^1|\!=\!|\mathcal{C}^2|\!=\ldots=\!|\mathcal{C}^{m-1}|$} while the base session with $\mathcal{D}^0$ has $C\!\gg\!N$ classes. We form $N$-way $K$-shot support set $\mathcal{S}^t$, 
\ie, an incremental session has $N$ classes and $K$ labeled nodes per class. The unlabeled nodes form the query set $\mathcal{Q}^t$. The model can access only $\mathcal{D}^t$ in session $t$ but it has to perform node classification on the union of subsets $\mathcal{D}^{\mathcal{T}} = \mathcal{D}^0 \cup  \mathcal{D}^1 \cup  ... \cup  \mathcal{D}^t$ over the encountered class space $\mathcal{C}^{\mathcal{T}} = \mathcal{C}^0 \cup  \mathcal{C}^1 \cup  ... \cup  \mathcal{C}^t$ observed up to point $t\!\leq\!m\!-\!1$.



\subsection{Prototype-based FSCIL Training Paradigm}


The prototype-based FSCIL uses class-wise prototypes $\boldsymbol{\rho}$ and the cosine similarity $sim(\cdot)$ to train the model and perform class predictions. Let $f_{\theta}(\cdot)$ be the feature extractor parameterized by $\theta$. 
Then, the prototype-based class prediction is given  by:
\begin{equation}
    p(y|\mathbf{x}) = \frac{e^{\tau (sim(\boldsymbol{\rho}_y, f_{\theta}(\mathbf{x})))}}{\sum_{y' \in \mathcal{C}^{\mathcal{T}}} e^{\tau(sim(\boldsymbol{\rho}_{y'}, f_{\theta}(\mathbf{x})))}},
\end{equation}

\noindent
where $\tau>0$ is the scaling factor. 
In the incremental session, $t$, the prototypes for newly occurred classes are calculated 
over the support set $\mathcal{S}^t$ of  $\mathcal{D}^t$:
\begin{equation}
    \boldsymbol{\rho}^t_c = \frac{\sum_{(\mathbf{x},y)\in\mathcal{S}^t} \mathbb{I}[y=c]f_{\theta}(\mathbf{x})}{\sum_{(\mathbf{x},y)\in\mathcal{S}^t} \mathbb{I}[y=c]},
    \label{eq:proto_calculate}
\end{equation}
where $\boldsymbol{\rho}^t_c$ is the prototype of class $c\!\in\!\mathcal{C}^t$ at session $t$, and $\mathbb{I}(\cdot)$ is the indicator function. 
The margin-based loss~\cite{wang2018cosface, zou2022margin, peng2022few} is used for training the model to obtain clear class boundaries:
\begin{equation}
\label{eq:margin_loss}
\!\!\!\!\!\mathcal{L}(\mathbf{x}, y) 
    = - log\, \frac{e^{\tau (sim(\boldsymbol{\rho}_y, f_{\theta}(\mathbf{x}))-\kappa )}}{e^{\tau (sim(\boldsymbol{\rho}_y, f_{\theta}(\mathbf{x}))-\kappa )} + \sum_{y' \neq y} e^{\tau (sim(\boldsymbol{\rho}_{y'}, f_{\theta}(\mathbf{x})))}},
\end{equation}
where 
$\kappa$ controls the class margin limiting 
 the variance of node embeddings concentrating around respective class prototypes. Intuitively, such margins help ``reserve'' space to accommodate novel classes.



\section{Methodology}
\begin{figure}
\vspace{-0.3cm}
    \centering
    \includegraphics[width=0.45\textwidth]{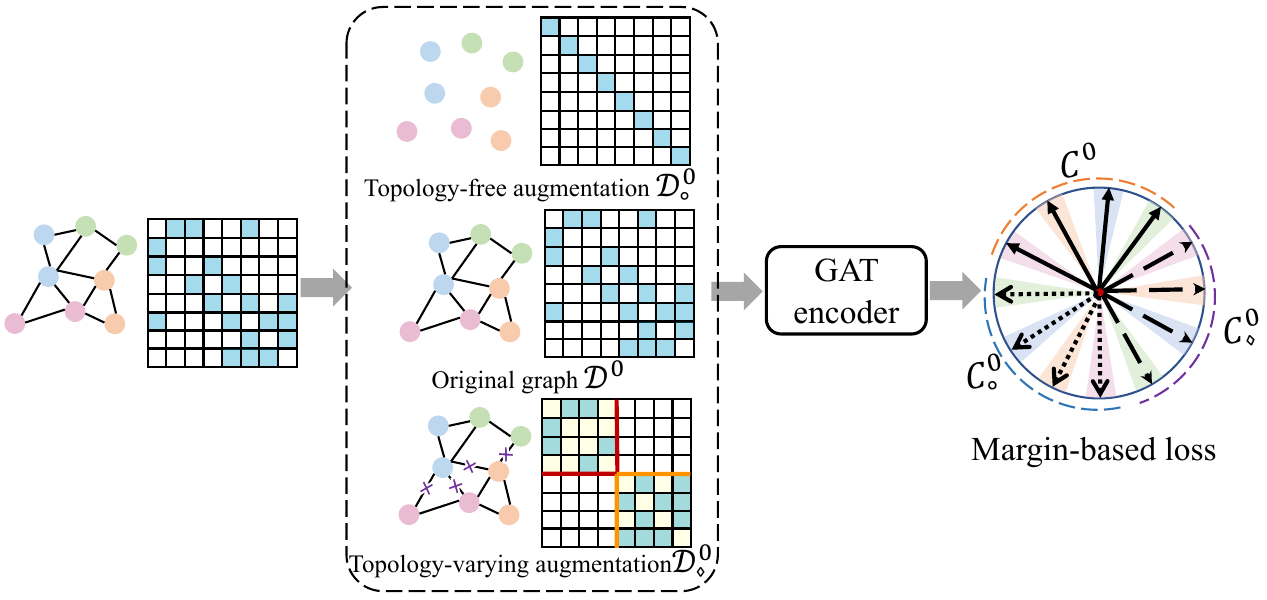}
    \vspace{-0.2cm}
   \caption{The triple-branch multi-topology class augmentation used during the base session training.}
    \label{fig:base_train}
    \vspace{-0.4cm}
\end{figure}

\begin{figure}
\vspace{-0.3cm}
    \centering
    \includegraphics[width=0.45\textwidth]{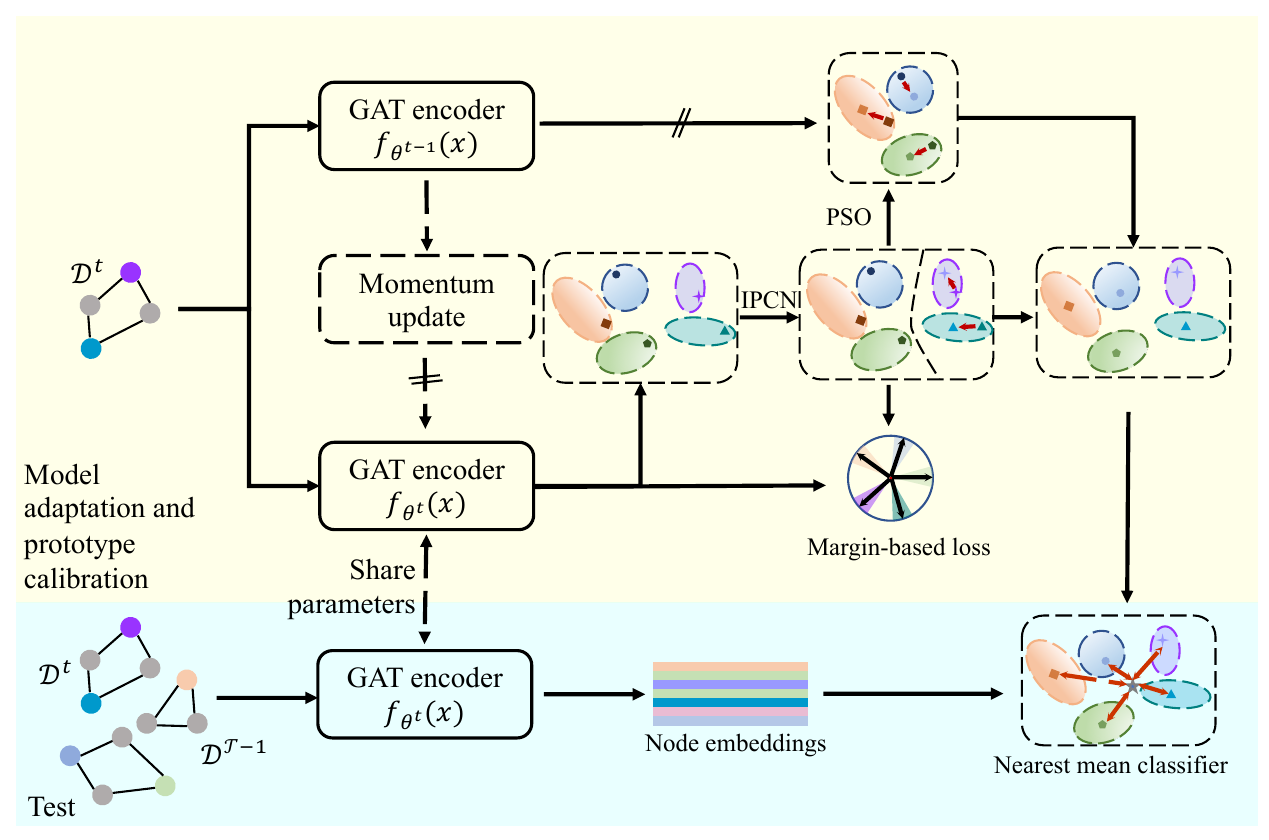}
    \vspace{-0.3cm}
    \caption{The training framework for model adaptation and prototype calibration in the incremental session.}
    \label{fig:novel_train}
    \vspace{-0.3cm}
\end{figure}

Our proposed TAP performs inductive GFSCIL. It is composed of three main components: (i) a Triple-branch Multi-topology Class Augmentation (TMCA), (ii) an Iterative Prototype Calibration for Novel classes (IPCN), and (iii) a Prototype Shift update for Old classes (PSO). Fig. \ref{fig:base_train} illustrates TMCA used for training during the base session with the goal of emulating the disjoint nature of incremental sessions, leading to a good pre-trained backbone. Fig.\ref{fig:novel_train} illustrates the incremental training framework that performs finetuning on new tasks. 
Concurrently, IPCN and PSO collectively contribute to calibrating and shifting the prototypes for novel and old classes, respectively. Algorithm \ref{alg:pseudo_code} details the training process. Below we introduce further technical details of our TAP. 

\subsection{Triple-branch Multi-topology Class Augmentation}
\label{sec:trple}
In inductive GFSCIL, each incremental session is likely to exhibit distinct structure patterns. Thus, we ease the dependency of the model on structural information 
by 
our proposed TMCA mechanism which uses (i) the entire base graph (full topology), (ii) graph nodes without links (topology-free), and (iii) subgraphs obtained by splitting base classes into subsets and then severing links between nodes belonging to different subsets (topology-varying). With emulations of several graph topologies, we assume that each topology should use its set of classes, \eg, we use $C$ class-prototypes per branch and separate $3C$ base classes, helping the backbone adapt to multiple topologies that the incremental sessions exhibit.

\vspace{0.1cm}
\noindent
\textbf{Topology-Free class Augmentation (TFA)} helps our model allocate a portion of its ``focus'' to semantic characteristics alone, helping class recognition independently of the graph structure. Specifically, TFA uses node features without any changes and eliminates all the graph links of base classes by substituting the original adjacent matrix with an identity matrix. Such an augmented graph is given its own label space. 
Formally, the augmented graph is represented as $\mathcal{D}^0_\circ=\{\mathcal{X}^0, \mathcal{A}^0_\circ, \mathcal{C}^0_\circ\}$ where $\mathcal{C}^0_\circ\!=\!\{c+|\mathcal{C}^0|\!: c\in \mathcal{C}^0\}$.


\vspace{0.1cm}
\noindent 
\textbf{Topology-Varying class Augmentation (TVA)}  enables the model to encounter a diverse array of structural patterns and teaches the model to extrapolate missing patterns beyond the base train set. TVA emulates the topology characteristics of novel classes in incremental sessions, preventing the model's overfitting to topology patterns of the base graph. We perform several TVA training epochs. At the beginning of each TVA epoch, base classes $\mathcal{C}^0$ are randomly split into $m'\!=\!\lceil C/N\rceil$ disjoint subsets $\big\{\mathcal{C}^{0}_{[0]}, \mathcal{C}^{0}_{[1]}, \ldots, \mathcal{C}^{0}_{[m'-1]}\big\}$, each comprising $N$ classes. Based on the class subsets, we also partition the graph adjacent matrix $\mathcal{A}^{0}$ into $\mathcal{A}^{0}_\diamond\!=\!\big\{\mathcal{A}^0_{[0]}, \mathcal{A}^0_{[1]}, \ldots, \mathcal{A}^{0}_{[m'-1]}\big\}$. 
Each partitioned adjacent matrix $\mathcal{A}^0_{[i]}$ is constrained to contain only links formed within the corresponding class subset $\mathcal{C}^0_{[i]}$. Thus, the links between subgraphs are removed. 
Also, a 10\% random link noise is injected into each $\mathcal{A}^{0}_{[i]}$ to increase the structural diversity of subsets. 
The total number of unique arrangements of $\mathcal{A}^0_{[i]}$ is:

\vspace{-0.2cm}
\begin{equation}
    \mathbb{C}_{|\mathcal{C}^0|}^{N} = \frac{|\mathcal{C}^0|!}{N!(|\mathcal{C}^0|-N)!},
\end{equation}
where $|\mathcal{C}^0|$ is the number of base classes. 

Such augmented subgraphs are given their own label space. Let $\Delta c\!=\!|\mathcal{C}^0|+|\mathcal{C}^0_\circ|$.  TVA-augmented  graph becomes $\mathcal{D}^{0}_\diamond=\{\mathcal{X}^0, \mathcal{A}^0_\diamond, \mathcal{C}^{0}_\diamond\}$ where $\mathcal{C}^{0}_\diamond\!=\!\big\{\big\{c\!+\!\Delta c\!: c\!\in\!\mathcal{C}^{0}_{[0]}\big\}, \ldots, \big\{c\!+\!\Delta c\!: c\!\in\!\mathcal{C}^{0}_{[m'-1]}\big\}\big\}$.


Given the generated class augmentations, the margin-based loss from Eq. \eqref{eq:margin_loss} is applied over both the original and augmented data during the base session training:
\begin{equation}
\mathcal{L}_B\!=\!\alpha\!\!\!\!\!\!\!\sum_{(\mathbf{x},y)\in \mathcal{D}^0}\!\!\frac{\mathcal{L}(\mathbf{x}, y)}{|\mathcal{D}^0|}+
     \alpha'\!\!\!\!\!\!\!\sum_{(\mathbf{x},{y})\in \mathcal{D}^0_\circ}\!\!\frac{\mathcal{L}({\mathbf{x}}, {y})}{|{\mathcal{D}^0_\circ}|} + \alpha'\!\!\!\!\!\!\!\sum_{({\mathbf{x}},{y})\in \mathcal{D}^0_\diamond}\!\!\frac{\mathcal{L}({\mathbf{x}}, {y})}{|\mathcal{D}^0_\diamond|},
\label{eq:loss_base}
\end{equation}
where $\alpha$ is the weight balancing the contribution between the original and the augmented graph data, and $\alpha'\!=\!(1\!-\!\alpha)/2$.

\subsection{Model Adaptation}
\label{sec:model_adapt}

Many approaches freeze their backbone and finetune the projection layer during incremental sessions \cite{zou2022margin, peng2022few}. However, this is insufficient for GFSCIL due to complex and polytropic structural information with distinctive topology patterns among subgraphs of base and novel classes. 

As inductive GFSCIL has to deal with novel structural patterns, we opt for updating the entire model in a momentum-based manner on the support set of incremental session $t$. 
Firstly, prototypes of novel classes $\mathcal{C}^t$ are computed according to Eq. \eqref{eq:pc-novel} (see Sec. \ref{sec:pcn} for details).  
The novel prototypes are then merged with all previous prototypes, denoted as $\mathcal{P}^{\mathcal{T}} = \mathcal{P}^{\mathcal{T}-1} 
\cup \{\boldsymbol{\rho}_{c}\!:c\!\in\!\mathcal{C}^t\}$. Subsequently, the encoder is trained on the support set $\mathcal{S}^t$ of $\mathcal{D}^t$ and the total label space $\mathcal{C}^{\mathcal{T}}$ over sessions $0,\ldots,t$ as:
\begin{equation}
    \mathcal{L}_N = \frac{1}{|\mathcal{S}^t|}\sum_{(\mathbf{x},y)\in \mathcal{S}^t} \mathcal{L}(\mathbf{x}, y).
    \label{eq:loss_novel}
\end{equation}

Note that as the number of labeled nodes of novel classes is low, causing overfitting, we perform only a few finetuning epochs. At the end of each session, we apply the Exponential Moving Average (EMA)  between past and current model parameters to prevent catastrophic forgetting. The final backbone parameters ${\theta}^{t}$ are 
    $\theta^t \leftarrow  \beta\,\theta^{t-1} + (1-\beta)\, \hat{\theta}^t,$ 
where $\theta^{t-1}$ and $\hat{\theta}^{t}$ are model parameters from session $t\!-\!1$ and session $t$ (after finetuning). 
$\beta$  controls EMA. 


\subsection{Prototype Calibration}
\subsubsection{Iterative Prototype Calibration for Novel classes (IPCN)}
\label{sec:pcn}
In the incremental session, due to the low number of nodes of the support set, the generated prototypes are not representative enough, leading to unsatisfactory performance. Thus, we propose an iterative prototype calibration, which uses the query set to calibrate novel prototypes iteratively (akin to k-means). Specifically, at each step of the incremental training, we use the latest encoder $f_{\hat{\theta}^t}(\cdot)$ to generate  probabilities of $\mathbf{x}_i$ belonging to $\boldsymbol{\rho}_{c}$ for the query nodes $\mathcal{Q}^t$ of novel classes:

\vspace{-0.6cm}
\begin{equation}
\label{eq:kmeans_pc}
p_c(\mathbf{x}_i) = \frac{e^{\tau\,sim(f_{\hat{\theta}^t}(\mathbf{x}_i), \boldsymbol{\rho}_c)}}{\sum_{c' \in \mathcal{C}^\mathcal{T}} e^{\tau\,sim(f_{\hat{\theta}^t}(\mathbf{x}_i), \boldsymbol{\rho}_{c'})}}.
\end{equation}
Probabilities $p_c(\mathbf{x}_i)$ are used to form pseudo-labeled query sets $\hat{\mathcal{Q}}^t_c$ used in the prototype calibration step: 
\begin{equation}
\begin{split}
    \hat{\boldsymbol{\rho}}^t_c = \frac{\sum_{\mathbf{x}_i \in \mathcal{S}^t_c} f_{\hat{\theta}^t}(\mathbf{x}_i) + \sum_{\mathbf{x}_j \in \hat{\mathcal{Q}}^t_c} p_c(\mathbf{x}_j) f_{\hat{\theta}^t}(\mathbf{x}_j)}{\sum_{\mathbf{x}_{i'} \in \mathcal{S}^t_c} 1 + \sum_{\mathbf{x}_{j'} \in \hat{\mathcal{Q}}^t_c} p_c(\mathbf{x}_{j'})},
\end{split}
\label{eq:pc-novel}
\end{equation}
where $\hat{\boldsymbol{\rho}}_c^t$ is the calibrated prototype for class $c$. 
The following steps are repeated few times: (i) Eq. \eqref{eq:kmeans_pc}, (ii) Eq. \eqref{eq:pc-novel}, (iii) $\boldsymbol{\rho}_c\!\leftarrow\!\hat{\boldsymbol{\rho}}_c^t,\forall c\in\mathcal{C}^t$. We perform 2 iterations of these steps.

\subsubsection{Prototype Shift for Old classes (PSO)}
Due to evolving parameter space from $f_{\theta^{t-1}}$ to $f_{\theta^t}$, the feature  distribution suffers from the drift. The prototypes of old classes do not represent old classes well when parameters $\theta^t$ change. Thus, we account for the drift: 
\begin{equation}
    \boldsymbol{\rho}^t_c = \boldsymbol{\rho}^{t-1}_c + \Delta \boldsymbol{\rho}_c^t,\forall c \in \mathcal{C}^{\mathcal{T}-1},
    \label{eq:pc-old}
\end{equation}

\noindent
where $\mathcal{C}^{\mathcal{T}-1}$ contains all classes encountered before   session $t$. $\Delta \boldsymbol{\rho}_c^t$ is the prototype drift vector which we estimate using the current support set $\mathcal{S}^t$ as the nodes of past sessions are unavailable.  
We assume the feature spaces of novel/previous tasks are not completely semantically disjoint. Thus, we approximate the prototype drift $\Delta \boldsymbol{\rho}_c^t$ by the feature changes 
%
$\Delta f_{\theta^{t-1} \rightarrow \theta^{t}}(\mathbf{x})\!=\!f_{\theta^t}(\mathbf{x}) - f_{\theta^{t-1}}(\mathbf{x}), \mathbf{x}\!\in\!\mathcal{S}^t$
%
and the corresponding empirical conditional density distribution $p(\boldsymbol{\rho}_c^{t-1}|\mathbf{x})$ \wrt $\boldsymbol{\rho}_c^{t-1}$:  
%
\begin{equation}
    \Delta \boldsymbol{\rho}^t_c = \sum_{\mathbf{x} \in \mathcal{S}^t} p(\boldsymbol{\rho}_c^{t-1}|\mathbf{x}) \Delta f_{\theta^{t-1} \rightarrow \theta^{t}}(\mathbf{x}).
\end{equation}
The empirical conditional density distribution $p(\boldsymbol{\rho}_c^{t-1}|\mathbf{x})$ \wrt $\boldsymbol{\rho}_c^{t-1}$ is given as:

\vspace{-0.6cm}
\begin{equation}
    p(\boldsymbol{\rho}_c^{t-1}|\mathbf{x}) = \frac{ \varphi(\mathbf{x}, \boldsymbol{\rho}^{t-1}_c)}{\sum_{\mathbf{x}' \in \mathcal{S}^t}\varphi(\mathbf{x}', \boldsymbol{\rho}^{t-1}_c)},
    \label{eq:dens}
\end{equation}
%
%
where $\varphi (\mathbf{x}, \boldsymbol{\rho}^{t-1}_c)\!=\!e^{-\frac{1}{2 \sigma ^2}\parallel f_{\theta^{t-1}}(\mathbf{x}) - \boldsymbol{\rho}^{t-1}_c \parallel^2_2}$ represents the Gaussian RBF similarity between the old prototype $\boldsymbol{\rho}^{t-1}_c$ and the node feature $f_{\theta^{t-1}}(\mathbf{x})$ generated by the previous encoder given a bandwidth hyperparameter $\sigma >0$. The PSO shifts the old prototypes along the direction of feature change in the neighborhood of prototypes. 



\section{Experiments}
\subsection{Experimental Setup}

\vspace{0.1cm}
\noindent
\textbf{Datasets.} 
To validate our proposed TAP method, we repurpose the existing common GNN datasets. 
As we have to perform incremental learning on several subgraphs with nodes of disjoint label space, we choose datasets featuring a substantial number of class categories, including Amazon\_clothing\cite{mcauley2015inferring}, DBLP\cite{tang2008arnetminer}, Cora\_full\cite{bojchevski2017deep}, and Ogbn-arxiv\cite{hu2020open}. The dataset statistics are summarized in Table \ref{table:dataset}.

\begin{algorithm}[t]
  \caption{Training with our TAP.}
  \label{alg:pseudo_code}
  \KwIn{A sequence of graph data subsets $\mathcal{D}=\{\mathcal{D}^0, \mathcal{D}^1, ..., \mathcal{D}^{m-1}\}$, GNN encoder $f_{\theta}(\cdot)$, hyperparameters $\alpha$, $\beta$, $\sigma$, $\tau, \kappa$. Epochs for base and incremental training $E_b, E_n$.}
  \KwOut{Predicted labels for the query nodes.}
  Initialized network parameters $\theta^0$.\\
  \tcc{\codebox{Base training session on $\mathcal{D}^0$.}}
  \For{$epoch=1,\ldots,E_b$}
  {
  Apply topology-free augmentation (TFA in Sec. \ref{sec:trple}).\\
  Get $m'\!=\!\lceil C/N\rceil$ topology-varying aug. partitions (TVA in Sec. \ref{sec:trple}).$\!\!\!\!$\\
  Update the GNN model $f_{\theta}(\cdot)$ according to Eq. \eqref{eq:loss_base}.\\
  } 
  \tcc{$\!\!\!$Incremental train. sessions on $\{\mathcal{D}^1,\ldots, \mathcal{D}^{m-1}\}\!\!\!$}
  \For{$t=1,\ldots,m-1$}
  {
  \tcc{\codebox{Model adaptation (incremental session $t$).$\!\!$}}
  \For{$epoch=1,\ldots,E_n$}
  {
  Generate node embeddings and class prototypes $\{\boldsymbol{\rho}_c\!:c\!\in\!\mathcal{C}^t\}$ for novel data $\mathcal{D}^t$ as Eq. \eqref{eq:proto_calculate}.\\
  Perform Iterative Prototype Calibration for Novel classes (IPCN): iterate 2x over (i) Eq. \eqref{eq:kmeans_pc}, (ii) Eq. \eqref{eq:pc-novel}, (iii) $\boldsymbol{\rho}_c\!\leftarrow\!\hat{\boldsymbol{\rho}}_c^t,\forall c\in\mathcal{C}^t$. \\
  Update the GNN model $f_{\theta^t}(\cdot)$ according to Eq. \eqref{eq:loss_novel}.\\
  }
  Apply EMA $\theta^t \leftarrow  \beta\,\theta^{t-1} + (1-\beta)\, \hat{\theta}^t$ from Sec. \ref{sec:model_adapt}.\\
  Prototype Shift for Old classes (PSO) as in Eq. \eqref{eq:pc-old}.\\
  \tcc{\codebox{Testing stage.}}
  Predict labels for all query nodes $\mathcal{Q}^\mathcal{T}$ in $\mathcal{D}^\mathcal{T}$ across all classes  observed to-date, given by $\mathcal{C}^\mathcal{T}$.
  }
\end{algorithm}

\begin{table}[t]
\vspace{-0.3cm}
    \centering
    \caption{Details of Four Benchmark Datasets.}
    \vspace{-0.2cm}
    \scalebox{0.8}{
    \begin{tabular}{ccccc}
         \toprule
         Dataset & Amazon\_clothing  & DBLP  & Cora\_full  & Ogbn-arxiv \\
         \midrule
          Nodes & 24,919 & 40,672 & 19,793 & 169,343  \\
          Edges & 91,680  & 288,270 & 65,311 & 1,116,243 \\
         Features & 9,034 & 7,202 & 8,710 & 128 \\
         Labels& 77 & 137 & 70 & 40  \\
         \midrule
         Base & 32 & 47 & 25 & 13  \\
         Novel & 45 & 90 & 45 & 27  \\
         N-way K-shot & (5, 5) & (10, 5) & (5, 5) & (3, 5) \\
         Sessions & 10 & 10 & 10 & 10\\
    \bottomrule
    \end{tabular}}
    \label{table:dataset}
    \vspace{-0.3cm}
\end{table}

Based on our problem formulation, we split datasets according to the base classes, $\mathcal{C}^0$, and novel classes, $\mathcal{C}^t, t>0$, subdivided into 9 incremental learning tasks, each formulated as $N$-way $K$-shot few-shot learning, as specified in Table \ref{table:dataset}.  Amazon\_clothing and Cora\_full are re-purposed into the 5-way 5-shot setting, Ogbn-arxiv into the 3-way 5-shot setting, while DBLP into the 10-way 5-shot setting. The remaining classes constitute the base classes, which are split into train (80\%) and test (20\%) sets.


We subsequently restructure the graph adjacency matrix into adjacency submatrices for incremental sessions. As nodes of subgraphs used in previous sessions are not supposed to be accessed in a new incremental session, we remove the links between subgraphs and retain graph links that reside within the same task. Thus, the graph structure of each task attains autonomy, devoid of any interdependence with other tasks (no overlap of node sets, link sets, or label spaces between tasks). 

\vspace{0.1cm}
\noindent
\textbf{Baseline methods.} Given the absence of existing methods tailored for the inductive GFSCIL, we adapt some representative methods from alternative domains as baselines, including CLOM \cite{zou2022margin}, LCWOF \cite{kukleva2021generalized}, SAVC \cite{song2023learning}, NC-FSCIL~\cite{yang2023neural}. We also adapt the \textit{transductive} GFSCIL method Geometer~\cite{lu2022geometer} to our setting. It should be noted that since Geometer~\cite{lu2022geometer} relies on access to the entire data graph (inclusive of the nodes of past tasks), we disable the component of past sample retraining for a fair comparison. 
What's more, a TAP variant and two widely-used FSCIL strategies have also been applied, establishing another three baselines as follows:
\renewcommand{\labelenumi}{\roman{enumi}.}
\begin{enumerate}[leftmargin=0.6cm]
    \item TAP-IPCN: It disables the prototype calibration for novel classes to evaluate the performance of TAP without the usage of unlabeled novel nodes. 
    \item GAT-finetune: It applies GAT as the backbone with the prototype-based classification method. In incremental learning, the entire network is finetuned using the support set of novel classes. 
    \item GAT-frozen: It also utilizes the GAT backbone. However, after the base training, the entire model is frozen and directly applied to query tests without further fine-tuning.
\end{enumerate}
To ensure a fair comparison, we apply the same GAT backbone and hyperparameters in all comparisons. 


\begin{table*}[t]
\vspace{-0.1cm}
    \centering
    \tabcolsep 3pt
    \caption{Overall performance comparison \wrt different sessions. Performance Drop (PD\textcolor{red}{$\downarrow$}) is measured between the base session and the last session (the smaller the drop the better). In each session, we evaluate also query nodes from former sessions. The improvement (Impr.\textcolor{green}{$\uparrow$}) is measured by the difference of the last session accuracies between TAP and other baselines.}
    \vspace{-0.4cm}
    \scalebox{0.95}{
        \begin{tabular}{lcccccccccccc|c}
             \toprule
              & \multicolumn{13}{c}{Amazon\_clothing (5-way 5-shot)} \\
             \cline{2-14} 
             Method & \multicolumn{13}{c}{Sessions} \\
             \cline{2-14} 
              &0& 1 &2 & 3 &4 &5 &6 & 7 &8 & 9 & A/Acc. & PD \textcolor{red}{$\downarrow$} & Impr. \textcolor{green}{$\uparrow$} \\
             \midrule
             GAT-finetune  &0.930 &0.121 &0.087 &0.103 &0.087 & 0.088 & 0.092 & 0.069 & 0.058 &0.064 & 0.170 & 0.867 & 0.604 \\
             GAT-frozen  &0.931 &0.861 &0.817 &0.756 &0.721 & 0.685 & 0.656 & 0.636 & 0.617 &0.591 & 0.727 & 0.340 & 0.077  \\             
             CLOM \cite{zou2022margin}  &\textbf{0.942} &0.879 &0.834 &0.775 &0.742 & 0.702 & 0.670 & 0.648 & 0.624 &0.595 & 0.742 & 0.343 & 0.073  \\
             LCWOF \cite{kukleva2021generalized}  &0.930 &0.863 &0.817 &0.755 &0.719 & 0.681 & 0.650 & 0.624 & 0.604 &0.578 & 0.722 & 0.352  & 0.090 \\
             SAVC \cite{song2023learning}  &0.934 &\underline{0.888} &\underline{0.850} &\underline{0.802} &\underline{0.766} & \underline{0.727} & \underline{0.693} & \underline{0.666} & \underline{0.643} &0.611 & \underline{0.759} & 0.325  & 0.057 \\
             NC-FSCIL~\cite{yang2023neural}  &0.840 &0.319 &0.442 &0.475 &0.492 & 0.449 & 0.480 & 0.414 & 0.405 &0.409 & 0.473 & 0.431  & 0.259 \\
             Geometer~\cite{lu2022geometer}  &0.934 &0.879 &0.835 &0.779 &0.743 & 0.705 & 0.682 & 0.662 & 0.639 &\underline{0.615} & 0.747 & \underline{0.319}  & 0.053 \\
             \midrule
             TAP-IPCN  &0.935&0.876&0.834&0.783&0.767&0.734&0.711&0.691&0.669&0.644&0.764&0.291 & 0.024 \\
             \rowcolor{LightCyan} \underline{TAP} &\underline{0.935} &\textbf{0.889} &\textbf{0.851} &\textbf{0.804} &\textbf{0.775} & \textbf{0.745} & \textbf{0.725} & \textbf{0.708} & \textbf{0.692} & \textbf{0.668} & \textbf{0.780} & \textbf{0.267}  & - \\
             \midrule 
             \midrule
              & \multicolumn{13}{c}{DBLP (10-way 5-shot)} \\
             \cline{2-14} 
             Method & \multicolumn{13}{c}{Sessions} \\
             \cline{2-14} 
              &0& 1 &2 & 3 &4 &5 &6 & 7 &8 & 9 & A/Acc.  & PD \textcolor{red}{$\downarrow$}  & Impr. \textcolor{green}{$\uparrow$}\\
             \midrule
             GAT-finetune  &0.637 &0.051 &0.054 &0.049 &0.044 & 0.038 & 0.036 & 0.035 & 0.031 & 0.033 & 0.101 & 0.603  & 0.470 \\
             GAT-frozen  &0.633 &0.605 &0.582 &0.559 &0.536 & 0.514 & 0.498 & 0.482 & 0.467 & 0.455 & 0.533 & 0.178  & 0.048 \\             
             CLOM \cite{zou2022margin}  &\textbf{0.644} &0.614 &0.589 &0.565 &0.541& 0.519 & 0.502 & 0.486 & 0.469 & 0.457 & 0.538 & 0.187  & 0.046 \\
             LCWOF \cite{kukleva2021generalized}  &0.631 &0.602 &0.578 &0.555 &0.532 & 0.509 & 0.493 & 0.477 & 0.461 &0.449 & 0.529 & 0.182   & 0.054\\
             SAVC \cite{song2023learning}  &0.624 &\underline{0.623} &\underline{0.611} &\underline{0.593} &\underline{0.574} & \underline{0.555} & \underline{0.539} & \underline{0.524} & \underline{0.507} &\underline{0.493} & \underline{0.564} & \underline{0.130}  & 0.010 \\
             NC-FSCIL~\cite{yang2023neural}  &0.427 &0.250 &0.237 &0.219 &0.207 & 0.183 & 0.183 & 0.169 & 0.167 &0.159 & 0.220 & 0.268  & 0.344 \\
             Geometer~\cite{lu2022geometer}  &0.640 &0.625 &0.599 &0.576 &0.553 & 0.529 & 0.510 & 0.495 & 0.478 &0.466 & 0.547 & 0.174  & 0.037 \\
             \midrule
             TAP-IPCN  &0.640 &0.624 &0.607 &0.589 &0.571 & 0.552 & 0.535 & 0.522 & 0.508 &0.497 & 0.564 & 0.140 & 0.006 \\
             \rowcolor{LightCyan} \textbf{TAP} &\underline{0.641} &\textbf{0.631} &\textbf{0.612} &\textbf{0.594} &\textbf{0.577} & \textbf{0.557} & \textbf{0.540} & \textbf{0.525} & \textbf{0.510} & \textbf{0.503} & \textbf{0.569} & \textbf{0.138}  & - \\
             \midrule     
             \midrule
              & \multicolumn{13}{c}{Cora\_full (5-way 5-shot)} \\
             \cline{2-14} 
             Method & \multicolumn{13}{c}{Sessions} \\
             \cline{2-14} 
              &0& 1 &2 & 3 &4 &5 &6 & 7 &8 & 9 & A/Acc.  & PD \textcolor{red}{$\downarrow$}  & Impr.  \textcolor{green}{$\uparrow$}\\
             \midrule
             GAT-finetune  &0.941 &0.121 &0.093 &0.089 &0.074 & 0.056 & 0.055 & 0.041 & 0.047 &0.049 & 0.157 & 0.892  & 0.466 \\
             GAT-frozen  &0.941 &0.823 &0.740 &0.675 &0.614 & 0.569 & 0.530 & 0.501 & 0.471 & 0.445 & 0.631 & 0.496  & 0.070 \\
             CLOM \cite{zou2022margin}  &\textbf{0.945} &\underline{0.845} &\underline{0.763} &\underline{0.700} &\underline{0.638} & \underline{0.592} & \underline{0.552} & \underline{0.522} & \underline{0.491} & \underline{0.464} & \underline{0.651} & \underline{0.481}  & 0.051\\
             LCWOF \cite{kukleva2021generalized}  &0.940 &0.814 &0.726 &0.662 &0.601 & 0.558 & 0.517 & 0.487 & 0.458 &0.431 & 0.619 & 0.509   & 0.084\\
             SAVC \cite{song2023learning}  &{0.942} &{0.835} &{0.755} &{0.687} &{0.625} & {0.579} & {0.539} & {0.509} & {0.478} &{0.450} & {0.640} & {0.492}  & 0.065 \\
             NC-FSCIL~\cite{yang2023neural}  &0.891 &0.623 &0.559 &0.499 &0.419 & 0.388 & 0.345 & 0.335 & 0.325 &0.291 & 0.468 & 0.599  & 0.224 \\
             Geometer~\cite{lu2022geometer}  &0.942 &0.835 &0.738 &0.677 &0.613 & 0.575 & 0.534 & 0.502 & 0.479 &0.447 & 0.634 & 0.499  & 0.068 \\
             \midrule
             TAP-IPCN  &0.943 &0.848 &0.771 &0.716 &0.658 & 0.618 & 0.579 & 0.548 & 0.524 &0.494 & 0.670 & 0.449 & 0.019 \\
             \rowcolor{LightCyan} \textbf{TAP} &\underline{0.944} &\textbf{0.853} &\textbf{0.782} &\textbf{0.725} &\textbf{0.673} & \textbf{0.634} & \textbf{0.598} & \textbf{0.570} & \textbf{0.541} & \textbf{0.515} & \textbf{0.683} & \textbf{0.429}  & - \\
             \midrule 
             \midrule
              & \multicolumn{13}{c}{Ogbn-arxiv (3-way 5-shot)} \\
             \cline{2-14} 
             Method & \multicolumn{13}{c}{Sessions} \\
             \cline{2-14} 
              &0& 1 &2 & 3 &4 &5 &6 & 7 &8 & 9 & A/Acc.  & PD \textcolor{red}{$\downarrow$}  & Impr. \textcolor{green}{$\uparrow$}\\
             \midrule
             GAT-finetune  & 0.625 &0.104 &0.085 &0.075 &0.082 &0.064 & 0.057 & 0.061 & 0.059 & 0.052 &0.126 & 0.573  & 0.336 \\
             GAT-frozen   & 0.640 &0.546 &0.509 &0.472 &0.454 &0.428 & 0.407 & 0.380 & 0.370 & 0.330 &0.454 & 0.310   & 0.058\\
             CLOM \cite{zou2022margin}  & 0.638 &0.574 &\underline{0.539} &0.476 &\underline{0.465} &0.429 & 0.409 & 0.379 & 0.367 & 0.342 &0.464 & 0.296  & 0.046 \\
             LCWOF \cite{kukleva2021generalized}  & 0.647 &\underline{0.579} &0.522 &\underline{0.481} &0.456 &\underline{0.444} & \underline{0.423} & \underline{0.395} & 0.383 & 0.351 &\underline{0.468} & \underline{0.296}  & 0.037 \\
             SAVC \cite{song2023learning}    &0.643 &0.550 &0.511 &0.475 &0.455 &0.431 & 0.413 & 0.392 & 0.386 & 0.347 &0.460 & 0.296  & 0.041 \\
             NC-FSCIL~\cite{yang2023neural}    &0.639 &0.145 &0.099 &0.099 &0.126 &0.151 & 0.160 & 0.161 & 0.150 & 0.159 &0.189 & 0.480  & 0.229 \\
             Geometer~\cite{lu2022geometer}  &\underline{0.651} &0.568 &0.520 &0.478 &0.456 & 0.437 & 0.424 & 0.394 & \underline{0.384} &\underline{0.352} & 0.466 & 0.299  & 0.036 \\
             \midrule
             TAP-IPCN  &0.669 &0.589 &0.555 &0.515 &0.481 & 0.452 & 0.440 & 0.407 & 0.396 &0.373 & 0.488 & 0.296 & 0.015 \\
             \rowcolor{LightCyan} \textbf{TAP} &\textbf{0.670} &\textbf{0.625} &\textbf{0.590} &\textbf{0.547} &\textbf{0.505}& \textbf{0.472} & \textbf{0.460} & \textbf{0.421} & \textbf{0.408} & \textbf{0.388} & \textbf{0.509} & \textbf{0.282}  & -\\
             \bottomrule 
        \end{tabular}}
    \label{table:overall}
    \vspace{-0.1cm}
\end{table*}

\vspace{0.1cm}
\noindent
\textbf{Implementation details.} 
Our TAP method uses a 2-layer GAT as the backbone with the hidden layer output of size 16, ReLU activation, weight decay of 0.0005, learning rate of 0.01, and dropout rate of 0.5. The number of attention heads is set to 12. We set $\alpha\!=\!0.7$. The momentum weight $\beta\!=\!0.95$. We set 5 steps for model finetuning. The iterations for IPCN are set as 2, and the bandwidth $\sigma\!=\!1.0$ for PSO. The scaling factor $\tau\!=\!15$ and margin $\kappa\!=\!0.1$  for the margin-based loss. 
Hyperparameters were selected on Cora\_full according to the last session accuracy. We withheld 30\% of query nodes for validation and used the remaining query nodes for testing. We did not tune hyperparameters on the remaining datasets.

\begin{table}[!htbp]
    \centering
    \tabcolsep 3pt
    \caption{Ablation study on various TAP variants.}
    \vspace{-0.3cm}
    \scalebox{0.7}{
        \begin{tabular}{lcccccccccc|c}
             \toprule
              & \multicolumn{11}{c}{Amazon\_clothing (5-way 5-shot)} \\
             \cline{2-12} 
             Method & \multicolumn{11}{c}{Sessions} \\
             \cline{2-12} 
              &0& 1 &2 & 3 &4 &5 &6 & 7 &8 & 9 & Impr. \textcolor{green}{$\uparrow$} \\
             \midrule
             GAT-frozen  &0.931&0.861&0.817&0.756&0.721&0.685&0.656&0.636&0.617&0.591&0.077 \\
             TAP-TMCA  &\textbf{0.944}&0.887&0.8510&0.797&0.765&0.740&0.716&0.698&0.672&0.650&0.018 \\
             TAP-TVA  &0.935&0.887&0.849&0.798&0.769&0.742&0.722&0.701&0.681&0.656&0.012 \\
             TAP-TFA  &0.933&0.879&0.842&0.791&0.763&0.729&0.706&0.685&0.667&0.654&0.014 \\
             TAP-IPCN  &0.935&0.876&0.834&0.783&0.767&0.734&0.711&0.691&0.669&0.644&0.024 \\
             TAP-PSO  &0.935&0.879&0.846&0.802&0.768&0.739&0.713&0.689&0.657&0.632&0.036 \\
             TAP-EMA  &0.934&0.667&0.377&0.271&0.223&0.196&0.171&0.143&0.132&0.106&0.562 \\
             TAP\_FProj  &0.896 &0.882 &0.842 &0.793 &0.764 & 0.729 & 0.709 & 0.687 & 0.667 &0.648 &0.020 \\
             \rowcolor{LightCyan} TAP &{0.935} &\textbf{0.889} &\textbf{0.851} &\textbf{0.804} &\textbf{0.775} & \textbf{0.745} & \textbf{0.725} & \textbf{0.708} & \textbf{0.692} & \textbf{0.668} &-  \\
             \bottomrule 
        \end{tabular}}
    \label{table:ablation}
\end{table}

\vspace{-0.3cm}
\subsection{Performance Comparisons}

Table \ref{table:overall} shows the experimental results for all the baselines over 10 sessions, in terms of their prediction accuracy at each session, the average accuracy (A/Acc.) across 10 sessions, and the corresponding performance drop (PD) between sessions 0 and 9. All the experiments are repeated 10 times with different random seeds. The results of the best-performing method are highlighted with \textbf{bold}, and the second-best baseline is underlined. 
Overall, TAP outperforms baselines by a large margin over all the datasets. Specifically, in the base session, most methods start from similar performances. Although the class augmentation method triples the label space for the base session, thereby increasing training complexity, TAP still manages to obtain outstanding results. 
With the progress of incremental sessions, TAP shows strengthening performance and exhibits growing improvements over other baselines. 
Concerning the last session accuracy, TAP achieves improvements of 5.7\%, 1.0\%, 6.5\%, and 3.1\% on Amazon\_clothing, DBLP, Cora\_full and Ogbn-arxiv datasets, respectively, compared with the best baselines. 
These advantages are further affirmed by the performance of A/Acc. and the PD across the four datasets. For instance, TAP has obtained a superior performance of 0.780 and 0.264 on Amazon\_clothing, compared with 0.759 and 0.319 of the best baseline in terms of A/Acc. and PD, respectively. 
In addition, despite minor declines relative to TAP, TAP-IPCN still surpasses all baselines with improvements of 2.9\%, 0.4\%, 3.0\%, 2.1\% on Amazon\_clothing, DBLP, Cora\_full and Ogbn-arxiv datasets, respectively, compared with the best baselines. This demonstrates that TAP can continue to achieve remarkable performance with minimal dependence on unlabeled nodes as incremental sessions progress.

Furthermore, a deeper examination of baseline performance reveals noteworthy insights. Firstly, GAT-frozen demonstrates a competitive performance among other baselines, supporting the effectiveness of the incremental-frozen strategy in preserving model capability as novel classes emerge. Secondly, NC-FSCIL performs even less favorably than GAT-frozen. This aligns with expectations, as solely fine-tuning the last projection layer instead of the feature extractor is ineffective in assimilating novel structural knowledge. GAT-finetune degrades recognition ability on old classes with a significant drop in the last session, diverging notably from the FSCIL methods observed in the image domain.

\vspace{-0.2cm}
\subsection{Validation on N-way one-shot Setting}
\vspace{-0.1cm}
\label{sec:1-shot}
\begin{figure*}
    \centering
    \includegraphics[trim={0 10 0 0},clip, width=0.9\textwidth]{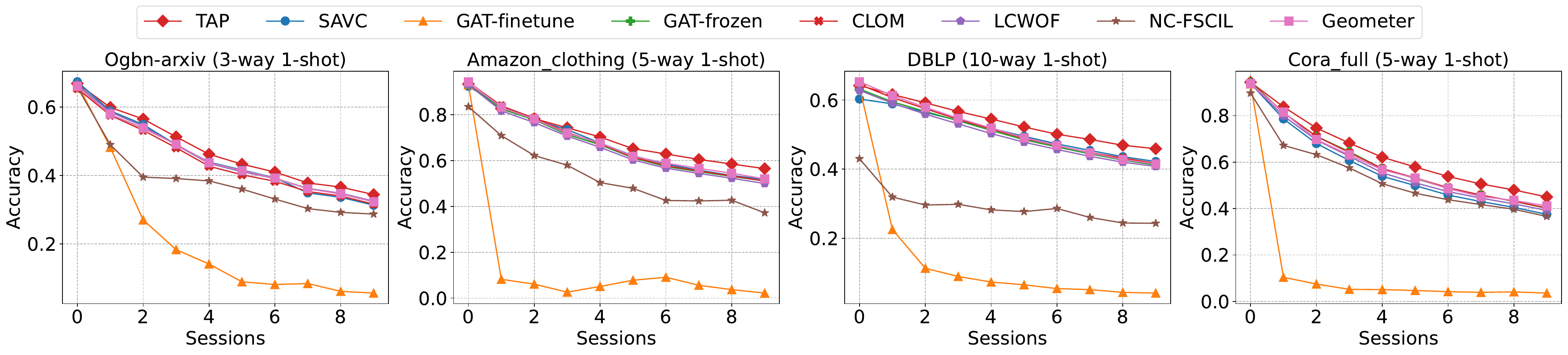}
    \vspace{-0.3cm}
    \caption{Performance comparison on the N-way 1-shot setting over different datasets.}
    \label{fig:1-shot}
\end{figure*}

To validate the performance of TAP under extreme label constraints, we present experimental results for the N-way 1-shot setting, including 3-way 1-shot on Ogbn-arxiv, 5-way 1-shot on Amazon\_clothing and Cora\_full, as well as 10-way 1-shot on DBLP. These results are summarized in Figure~\ref{fig:1-shot}, with specific details provided in Appendix \ref{appendix:1-shot}. Overall, TAP consistently outperforms all other methods across the datasets, with its performance curve distinctly higher than the others. It demonstrates the slowest performance decline as incremental learning and maintaining the highest accuracy in the last session. In contrast, GAT-finetune and NC-FSCIL perform the worst, likely due to their inability to address the forgetting problem in incremental training. The rest perform at a similar level, but none can match TAP's performance. Notably, TAP’s advantages become more pronounced in the later sessions, benefiting from prototype calibration, which mitigates the issue of overfitting and forgetting. The performance aligns with that observed in Table \ref{table:overall}, further highlighting TAP's strong robustness against the overfitting risks associated with extreme label sparsity.

\subsection{Ablation Study}


Below we ascertain the efficacy of individual components of TAP by systematically removing each component, whose results are presented in Table \ref{table:ablation}. The notation "TAP-*" refers to the ablation where the component "*" is removed from TAP, as specified below:
\renewcommand{\labelenumi}{\roman{enumi}.}
\begin{enumerate}[leftmargin=0.6cm]
    \item TAP-TMCA: It removes the whole class augmentation method, including TFA and TVA.
    \item TAP-TFA: It removes the topology-free augmentation method.
    \item TAP-TVA: It removes the topology-varying augmentation method.
    \item TAP-PSO: It removes the prototype shift for old classes.
    \item TAP-IPCN: It removes the prototype calibration for new classes.
    \item TAP-CE: It replaces the margin-based loss with the cross-entropy loss, equivalent to specifying  $\kappa\!=\!0$ for margin.
    \item TAP-EMA: It removes the exponential moving average operation between the novel and old models. 
    \item TAP\_FProj: It refers to the model where the GAT backbone is frozen, and only an appended projection layer is fine-tuned.
\end{enumerate}
When comparing the last session accuracy with TAP, we observe that removing TMCA (i.e., TVA + TFA), TVA, and TFA results in a performance drop of 1.8\%, 1.2\%, and 1.4\%, respectively. The removal of IPCN and PSO, which disables prototype calibration for novel and old classes, leads to performance degradations of 2.4\% and 3.6\%, respectively. 
This highlights the importance of prototype calibration as incremental learning progresses. The utility of EMA is to prevent the model from catastrophically forgetting previously obtained knowledge after model fine-tuning. Thus removing EMA can lead to a collapsed model. Additionally, the performance of TAP\_FProj also drops significantly (0.648 vs. 0.668), underscoring the need to update the entire model, rather than just the appended projection layer, to accommodate the evolving graph structure—unlike FSCIL in the image domain, where structural information is absent. 
Furthermore, we also provide extra ablation studies on multiple \textbf{\textit{class augmentation variants}}, please refer to Appendix \ref{appendix:augment} for more details.

\vspace{-0.2cm}
\subsection{Visualization of Prototype Calibration}



\begin{figure}
\vspace{-0.3cm}
    \centering
     \subfigure[PSO]{
         \includegraphics[width=0.22\textwidth]{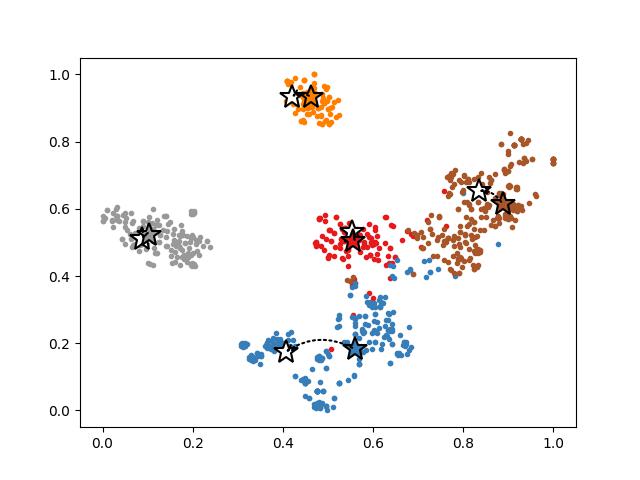}
         \label{fig:pco}}
     \subfigure[IPCN]{
         \includegraphics[width=0.22\textwidth]{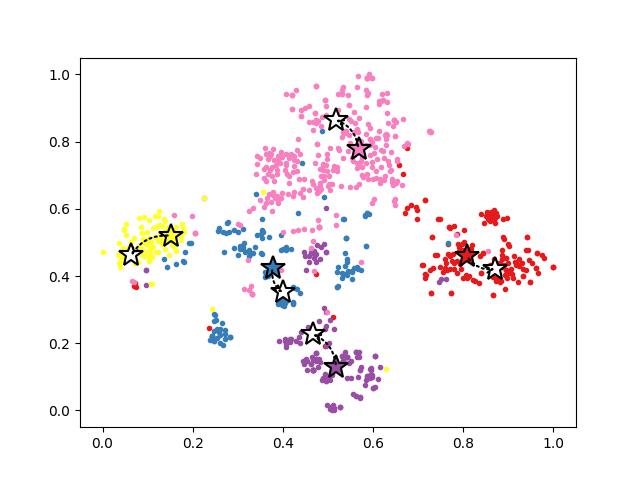}
         \label{fig:pcn}}
         \vspace{-0.3cm}
    \caption{The visualization of prototype calibration for old and novel classes on Amazon\_clothing, where the empty and colored stars indicate prototypes before and after calibration. The dotted lines denote trajectories of prototype movements.}
    \label{fig:proto_calib}
\end{figure}

Fig. \ref{fig:proto_calib}  visualizes prototype calibration for old and novel classes, respectively. The empty and colored stars indicate the prototypes before and after calibration, and their moving trajectories are depicted by dotted lines. 
Fig. \ref{fig:pco} presents the calibrating result of PSO for 5 base classes, illustrating the refinement of stored prototypes through estimated prototype shift. Notably, most original prototypes initially reside in the marginal areas of their respective classes, exhibiting substantial drifts from their class centroids. The PSO pulls the prototypes towards denser areas, generating more representative prototypes. 
Fig. \ref{fig:pcn} shows the prototype adjustment of 5 novel classes induced by IPCN, which relocates those prototypes closer to their class centroids, reducing position variability within their class communities, and thus improving their representativeness. 

\vspace{-0.2cm}
\subsection{Parameter Analysis}

\begin{figure}
\vspace{-0.4cm}
    \centering
    \subfigure[$\alpha$]{
    \includegraphics[width=0.215\textwidth]{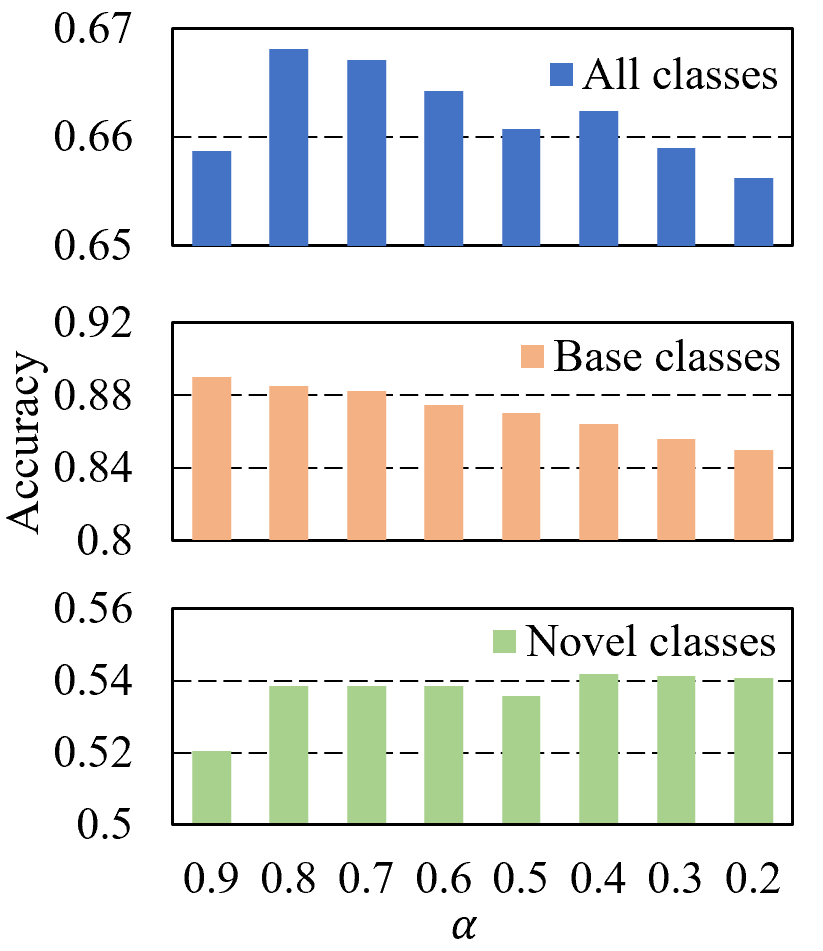}
    \label{fig:alpha}
    }
    \subfigure[$\beta$]{
    \includegraphics[width=0.22\textwidth]{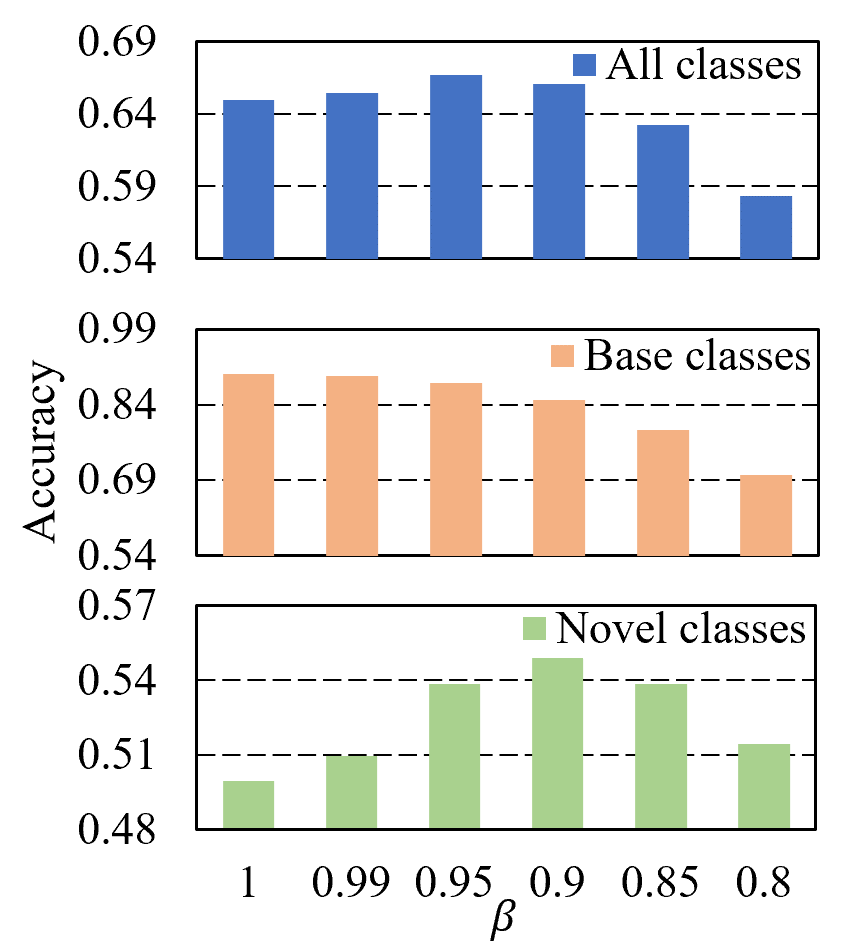}
    \label{fig:beta}
    }
    \vspace{-0.4cm}
    \caption{The impact of $\alpha$ and $\beta$ on the last session accuracy on Amazon\_clothing over all, base and novel classes. }
    \label{fig:sigma}
    \vspace{-0.4cm}
\end{figure}

\vspace{-0.1cm}
\noindent
\textbf{Impact of $\alpha$.}  The parameter $\alpha$ regulates the influence of the topology-free and topology-varying class augmentations as described in Eq. \eqref{eq:loss_base}. A larger value of $\alpha$ indicates a more dominant role of augmented data in base training. 
Fig.~\ref{fig:alpha} presents the sensitivity analysis results of $\alpha$ on Amazon\_clothing dataset. The three histograms illustrate the accuracy of the last session across all encountered classes, base classes, and novel classes, respectively. The overall performance on all classes indicates a preference for a relatively larger value of $\alpha$ as opposed to smaller ones. The method achieves its peak performance when $\alpha\!\in\![0.7, 0.8]$, gradually diminishing as $\alpha$ deviates away from this range.  
Examining the performance changes on base and novel classes, a decline in $\alpha$ from 0.9 to 0.2 results in a continuous performance degradation in base classes. Conversely, novel classes initially experience performance rises, followed by stabilization in the later period. This observation underscores the efficacy of our class augmentation in enhancing the model's generalizability to unseen classes and structural patterns.

\vspace{0.1cm}
\noindent
\textbf{Impact of $\beta$.} 
%
%
The parameter 
$\beta$ governs the momentum-based balance between the contribution of the previous and the latest model parameters. 
The sensitivity analysis of $\beta$ in Fig.\ref{fig:beta} shows that TAP achieves optimal performance for $\beta\!\in\![0.9, 0.95]$, but experiences performance decline outside of this range. Specifically, for $\beta>0.95$, the model struggles to assimilate sufficient novel knowledge from model fine-tuning, resulting in unsatisfactory improvement, as evidenced by declining performance on novel classes. Conversely, when $\beta<0.9$, the model suffers from catastrophic forgetting, leading to a significant performance drop on the base classes.

\begin{figure}
\vspace{-0.3cm}
    \centering
    \subfigure[]{
    \includegraphics[trim={0 15 0 0},clip,width=0.22\textwidth]{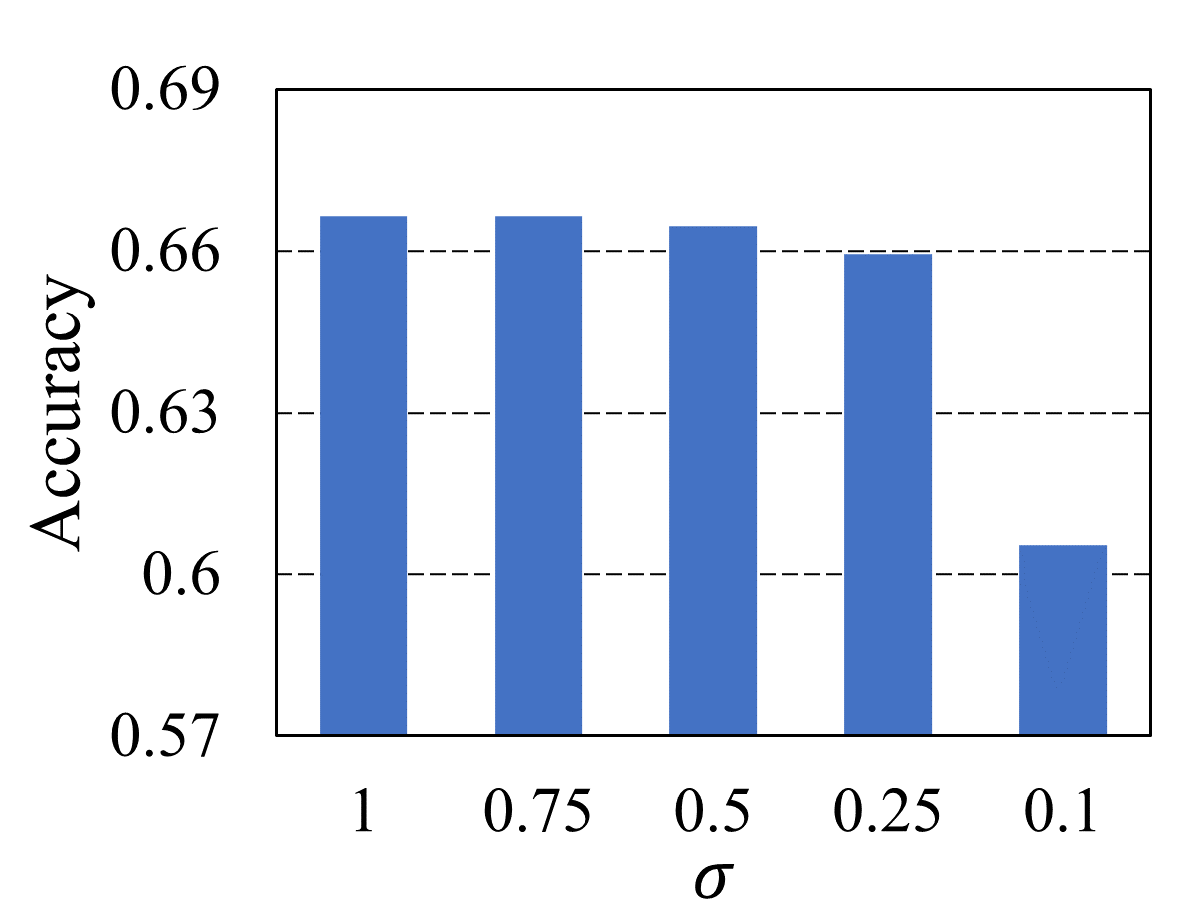}
    \label{fig:sigma_acc}
    }
    \subfigure[]{
    \includegraphics[trim={0 0 0 0},clip,width=0.21\textwidth]{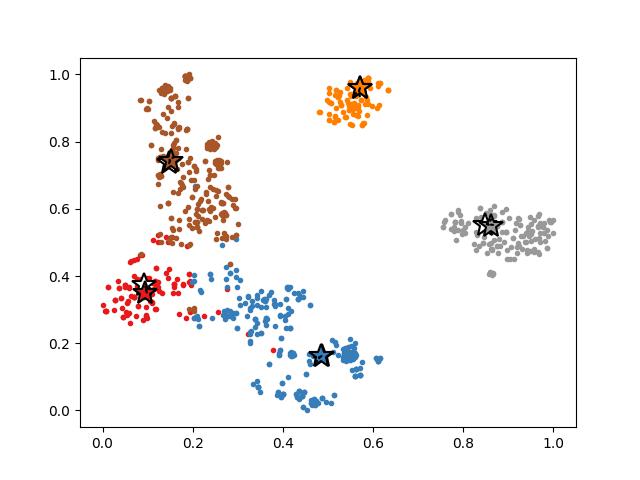}
    \label{fig:tsne-pco-sigma}
    }
    \vspace{-0.3cm}
    \caption{The impact of $\sigma$ on the model performance over the Amazon\_clothing dataset. Fig. \ref{fig:sigma_acc}: The impact of $\sigma$ on the accuracy of the last session. FIg. \ref{fig:tsne-pco-sigma}: The prototype change on old classes when $\sigma\!=\!0.25$.}
    \label{fig:sigma}
    \vspace{-0.4cm}
\end{figure}

\vspace{0.1cm}
\noindent
\textbf{Impact of $\sigma$.} The parameter $\sigma$ denotes the bandwidth as described below Eq. \eqref{eq:dens}. It governs the weight distribution for the prototype shift estimation. 
Fig.\ref{fig:sigma_acc} shows that a decrease in $\sigma$ from 1.0 to 0.5 does not cause notable performance change. 
However, once it descends to 0.25 or less, the performance drops as for small $\sigma$, all weights tend to approach zero, rendering the ineffective calibration. This is evident in Fig. \ref{fig:tsne-pco-sigma}, where the adjusted prototypes stay close to the original prototypes after calibration. 


\vspace{-0.2cm}
\section{Conclusions}
In this paper, the inductive GFSCIL problem is formulated and effectively addressed with our proposed TAP model. The primary challenge of inductive GFSCIL  stems from issues of catastrophic forgetting and overfitting, arising due to the unavailability of data from previous sessions and the extreme scarcity of labeled samples.  
To this end, we have introduced topology-free and topology-varying class augmentations for the base training, emulating the disjoint topology of subgraphs arriving in incremental sessions. We have proposed an iterative prototype calibration for refining novel prototypes during incremental training to reduce their variance and enhance their representativeness. 
Furthermore, to address the inconsistency between node embeddings and stored old prototypes resulting from model fine-tuning, we have introduced a prototype shift to recalibrate the old prototypes, effectively accounting for the feature distribution drift. Finally, we have substantiated the effectiveness of our TAP through extensive experiments, comparing its performance to state-of-the-art baselines across four datasets. 

\begin{acks}

This work was partially funded by CSIRO's Reinvent Science and CSIRO's Data61 Science Digital. The authors gratefully acknowledge continued support from the CSIRO's Data61 Embodied AI Cluster. This work was also supported in part by the 2024 Gansu Province Key Talent Program under Grant No.2024RCXM22. 
    
\end{acks}




\bibliographystyle{ACM-Reference-Format}
\bibliography{sample-base}

\clearpage
\appendix



\end{document}